\documentclass{article}

\usepackage[final]{corl_2018} %

\usepackage{graphicx}
\usepackage{amssymb}
\usepackage{amsmath}
\usepackage{authblk}
\usepackage{mathtools}
\usepackage{booktabs}
\usepackage{pifont}
\usepackage{subcaption}

\newcommand{\cmark}{\ding{51}}%

\usepackage{caption}
\graphicspath{ {images/} }

\usepackage{floatrow}
\newfloatcommand{capbtabbox}{table}[][\FBwidth]
\usepackage{blindtext}

\newcommand\newsubcap[1]{\phantomcaption%
       \caption*{\figurename~\thefigure\thesubfigure: #1}}

\title{IntentNet: Learning to Predict Intention \\ from Raw Sensor Data}

\author{
\textbf{Sergio Casas, Wenjie Luo, Raquel Urtasun} \\
Uber Advanced Technologies Group, University of Toronto \\
\texttt {\{sergio.casas,wenjie,urtasun\}@uber.com}
}

\begin{document}
\maketitle

\begin{abstract}
In order to plan a safe maneuver, self-driving vehicles need to understand the intent of  other traffic participants. We define intent as a combination of discrete high level behaviors as well as continuous trajectories describing future  motion. 
In this paper we  develop a one-stage detector and forecaster that exploits  both 3D point clouds produced by a LiDAR sensor as well as dynamic maps of the environment. 
Our multi-task model achieves better accuracy than the respective separate modules while saving computation,  which is critical to reduce reaction time in self-driving applications. 
\end{abstract}


\section{Introduction}

Autonomous driving is one of the most exciting problems of modern artificial intelligence. Self-driving vehicles have the potential to revolutionize the way  people and freight move.  While a plethora of  systems have been built in the past few decades, many challenges still remain. One of the fundamental difficulties is that self driving vehicles have  to share the roads with human drivers, which can perform maneuvers that are  difficult to predict. 

Human drivers  understand intention by exploiting the actors'  past motion  as well as     prior knowledge about the scene (e.g. the location of lanes, direction of driving). %
Previous work \cite{streubel2014prediction, phillips2017generalizable, hu2018probabilistic}  attempted to solve this problem by first performing vehicle detection and then extracting intent from the position and motion of detected bounding boxes. 
This, however, restricts the information that the intent estimation module has access to, resulting in suboptimal estimates. 
Very recently, FaF \cite{faf} directly exploited  LiDAR sensor data to predict the  future trajectories of vehicles. However, the trajectories were only predicted for 1 second into the future and no intent prediction was done beyond motion estimation.

In this paper, we take this approach one step further and propose a novel deep neural network that reasons about both high level behavior and long term trajectories. Inspired by how humans perform this task, we design a network that exploits motion and prior knowledge about the road topology in the form of maps containing semantic elements such as  lanes, intersections and traffic lights.
In particular, our {\it IntentNet} is a fully-convolutional neural network that outputs three types of variables in a single forward pass corresponding to: detection scores for vehicle and background classes, high level action probabilities corresponding to discrete intention, and bounding box regressions in the current and future time steps to represent the intended trajectory. Our architecture allows us to jointly optimize all tasks, fixing the distribution mismatch problem between tasks when solving them separately. Importantly, our design also enables the system to propagate uncertainty through the different components. %
In addition, our approach is computationally efficient by design as all tasks share the heavy neural network feature computation.

We demonstrate the effectiveness of our approach in the tasks of detection and intent prediction by showing that our system surpasses other real-time, state-of-the art detectors  while outperforming previous  intent prediction approaches, both in its continuous and discrete counterparts.
In the remainder of the paper, we first discuss related work and then present our model followed by experimental evaluation and conclusion.

\section{Related Work} \label{related}

In this section we first review recent advances in object detection,  focusing on single-stage and 3D object detection. We then discuss  motion and intent prediction approaches. %

\paragraph{Object detection:}
Many proposal based approaches \cite{ren2015faster, dai2016r} have been developed after the seminal work of RCNN \cite{girshick2014rich}. While these methods perform really well, they are typically not suitable for real-time applications as they are  computationally demanding. In contrast, single stage detectors  \cite{redmon2016you,liu2016ssd} provide a more efficient solution. 
YOLO  \cite{redmon2016you} breaks down the image into grids and makes multi-class and multi-scale predictions at each cell. SSD \cite{liu2016ssd} added the notion of anchor boxes, which reduces object variance in size and pose. RetinaNet \cite{lin2017focal} showed that single-stage detectors can outperform two-stage detectors in both speed and accuracy. Geiger \textit{et al.} \cite{geiger2011joint} improved the ability to estimate object orientation by jointly reasoning about the scene layout.  More recently, Vote3Deep \cite{engelcke2017vote3deep} proposed to voxelize point clouds and exploit 3D CNNs \cite{tran2015learning}. Subsequently, FaF \cite{faf} and PIXOR \cite{pixor} exhibited superior performance in terms of speed and accuracy by exploiting a bird's eye view representation. Additionally,  \cite{faf}  aggregated   several point clouds from the past. Approaches that use the projection representation (VeloFCN \cite{li2016vehicle}, MV3D \cite{chen2017multi}) or handle point clouds directly (PointNet \cite{qi2017pointnet}) have also been proposed. However, these methods suffer from either limited performance or heavy computation and thus are not suitable for self driving \cite{simon2018complex}.  Liang \textit{et al.}  \cite{liang2018deep} proposed a real time 3D detector that exploits  multiple sensors (i.e., camera and LiDAR). 

\paragraph{Motion Forecasting: }
This refers to the task of predicting future locations of an actor given current and past information. 
DESIRE \cite{lee2017desire} introduced an RNN encoder-decoder framework in the setting of multiple interacting agents in dynamic scenes. Ma \textit{et al.} \cite{ma2017forecasting} proposed to couple game theory and deep learning to model pedestrians interactions and estimate person-specific behavior parameters. Ballan \textit{et al.} \cite{ballan2016knowledge} exploited the interplay between the dynamics of moving agents and the semantics of the scene for scene-specific motion prediction. Soo \textit{et al.} \cite{soo2016egocentric} created an EgoRetinal map to predict plausible future ego-motion trajectories  in egocentric stereo images. Hoermann {et al.} \cite{hoermann2017dynamic} utilized a dynamic occupancy grid map as input to a deep convolutional neural network to perform long-term situation prediction in autonomous driving. SIMP \cite{hu2018probabilistic}  parametrized the output space as insertion areas where the  vehicle  of interest could go, predicting  an estimated time of arrival and a spatial offset. Djuric \textit{et al.} \cite{djuric2018motion}  rasterized representations of each actor's vicinity in order to predict their future motion. FaF \cite{faf} pioneered the unification of  detection and short term motion forecasting from LiDAR point clouds in driving scenarios. 

\paragraph{Intention Prediction:}
The intention of an actor can be seen as the  sequence of actions it will take in order to achieve an objective. Fathi \textit{et al.} \cite{fathi2012learning}  developed a probabilistic generative model to predict daily actions using gaze. In the context of intelligent vehicles, Zhang \textit{et al.} \cite{zhang2013understanding} proposed to model high level semantics in the form of traffic patterns through a generative model that also reasons about the geometry and objects present in the scene. Jain \textit{et al.} \cite{jain2015car} proposed an autoregressive HMM to anticipate driving maneuvers a few seconds before they occur by exploiting video from a face and a rear camera together with features from the map. Streubel \textit{et al.} \cite{streubel2014prediction} utilized HMMs to estimate the direction of travel while approaching a 4-way intersection. Kim \textit{et al.} \cite{kim2017prediction} predicted egocentric intention of lane changes. Recently, Phillips \textit{et al.}  \cite{phillips2017generalizable} utilized LSTMs to predict cars' intention at generalizable intersections. SIMP \cite{hu2018probabilistic} suggested to model intention implicitly by defining a discrete set of insertion areas belonging to particular lanes. Unfortunately, most of the work in this area lacks solid evaluation. For instance, \cite{phillips2017generalizable} used 1 hour of driving data across only 9 intersections for both training and evaluation (cross-validated), while \cite{hu2018probabilistic} apply the SIMP framework to only 640 meters of highway. \\

IntentNet is inspired by FaF \cite{faf}, which performs joint detection and future prediction. IntentNet achieves a more accurate vehicle detection and trajectory forecasting, enlarges the prediction horizon and estimates future high-level driver's behavior. The key contributions to the performance gain are (i) a more suitable architecture based on an early fusion of a larger number of previous LiDAR sweeps, (ii) a parametrization of the   map that allows our model to understand traffic constraints for all vehicles at once and (iii) an improved loss function that includes a temporal discount factor to account for the inherent ambiguity of the future.

\section{Learning to Predict Intention}
\label{sec:method}

In this section, we present  our approach to  jointly detect vehicles and  predict  their  intention   directly from raw sensor data. Towards this goal, we exploit  3D LiDAR point clouds and dynamic HD  maps containing semantic elements such as  lanes, intersections and  traffic lights.
In the following,  we describe  our  parametrization, network architecture, learning objective and inference procedure.

\subsection{Input parametrization} \label{input_param}

\paragraph{3D point cloud:}
Standard convolutional neural networks (CNNs) perform discrete convolutions, assuming a grid structured input. Following \cite{chen2017multi, faf, pixor}, we represent point clouds in bird's eye view (BEV) as a 3D tensor, treating height as our channel dimension.
This input parametrization has several key advantages: (i) computation efficiency due to dimensionality reduction (made possible as vehicles drive on the ground), (ii) non-overlapping targets (contrary to camera-view representations, where objects can overlap),  (iii) preservation of the metric space (undistorted view) that eases the creation of priors regarding vehicle sizes, and
 (iv)  this representation also makes the fusion of LiDAR and map features trivial as both are defined in bird's eye view.
We utilize multiple consecutive LiDAR sweeps (corrected by ego-motion) as the past is fundamental to accurately estimate both intention and motion forecasting.
We diverge from previous work and stack together height and time dimensions into the channel dimension as this allows us to use 2D convolutions to fuse time information. As shown in our experiments, this is  more effective than the non-padded 3D convolutions   proposed in \cite{faf}. %
This gives us a tensor of size: $(\frac{L}{\Delta L}, \frac{W}{\Delta W}, \frac{H}{\Delta H} \cdot T)$, where $L$, $W$ and $H$ are the longitudinal, transversal and normal physical dimensions of the scene; $\Delta L$, $\Delta W$ and $\Delta H$ are the voxel sizes in the corresponding directions and $T$ is the number of LiDAR sweeps.

\begin{figure}[t]
\centering
  	\begin{subfigure}[t]{\textwidth}
    	\centering
    	\includegraphics[width=\linewidth]{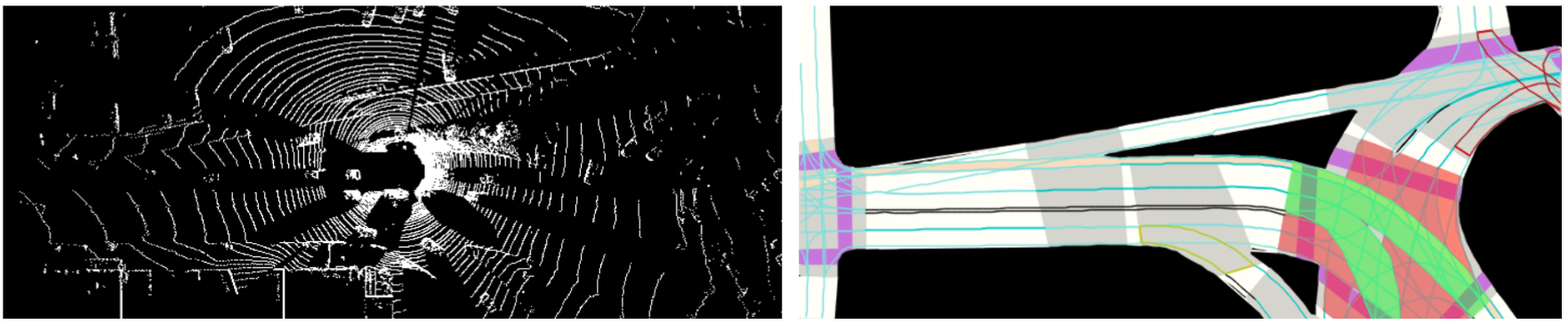}
  	\end{subfigure}
\caption{Input parametrization. Left: Voxelized LiDAR in BEV (height aggregated in a single channel for visualization purposes). Right: Rasterized map (in RGB for visualization purposes)}
\label{fig:input_param}
\end{figure}

\paragraph{Dynamic maps:} We form a BEV representation of our maps by  rasterization. We exploit     static information including roads, lanes, intersections, crossings and traffic signs, in conjunction with traffic lights, which contain  dynamic information that changes over time (i.e., traffic light state changes between green, yellow and red).
We represent each semantic component in the map with  a binary map (i.e., 1 or -1).
Roads and intersections are represented as filled polygons covering the whole drivable surface. Lane boundaries are parametrized as poly-lines representing the left and right boundaries of lane segments. Note that  we use three binary masks to distinguish lane types, as lane boundaries  can be crossed or not, or only in certain situations. Lane surfaces are rasterized to distinguish between straight, left and right turns, as this information is helpful for intention prediction. We also use two extra binary masks for bike and bus lanes as a way to input a prior of non-drivable road areas.
Furthermore, traffic lights can change the drivable region dynamically. We encode the state of the traffic light into the lanes they govern. We rasterize the surface of the lane succeeding the traffic light in one out of three binary masks depending on its state: green, yellow or red.
One extra layer is used to indicate whether those lanes are protected by its governing traffic light, i.e. cars in other lanes must yield. This situation happens in turns when the arrow section of the traffic light is illuminated.
We estimate the traffic light states using cameras in our self-driving vehicle.
We also infer the state of some unobserved traffic lights that directly interact with known traffic light states. For example, a straight lane with unknown traffic light state that collides with a protected turn with green traffic light state can be safely classified as being red.
Lastly, traffic signs are also encoded into their governed lane segments, using two binary masks to distinguish between yields and stops.
In total, there are 17 binary masks used as map features, resulting in a 3D tensor that represents the map.
Fig.  \ref{fig:input_param} shows an example, where  different elements (e.g., lane markers in cyan, crossings in magenta, alpha blended traffic lights with their state colored) are depicted.

\subsection{Output parametrization} \label{output_param}

Our model predicts  drivers' intentions in both  discrete and continuous form.

\paragraph{Trajectory regression:}
For each detected vehicle, we parametrize its  trajectory as a sequence of bounding boxes, including current and future locations. Assuming cars are non-deformable objects, we treat their size ($w$, $h$) as a constant estimated by the detector. The pose in each time stamp is  3D and contains the bounding box center ($c_x^t$, $c_y^t$) and heading  $\phi^t$ of the vehicle in BEV coordinates (see Fig. \ref{fig:output_param}).

\paragraph{High level actions:}

We frame the discrete intention prediction problem as a multi-class classification with 8 classes: \textit{keep lane, turn left, turn right, left change lane, right change lane, stopping/stopped, parked and other}, where \textit{other} can be any other action such as reversed driving.  %

\subsection{Network architecture} \label{architecture}
\begin{figure}[t]
\begin{subfigure}[t]{.34\textwidth}
    \centering
    \includegraphics[width=\linewidth]{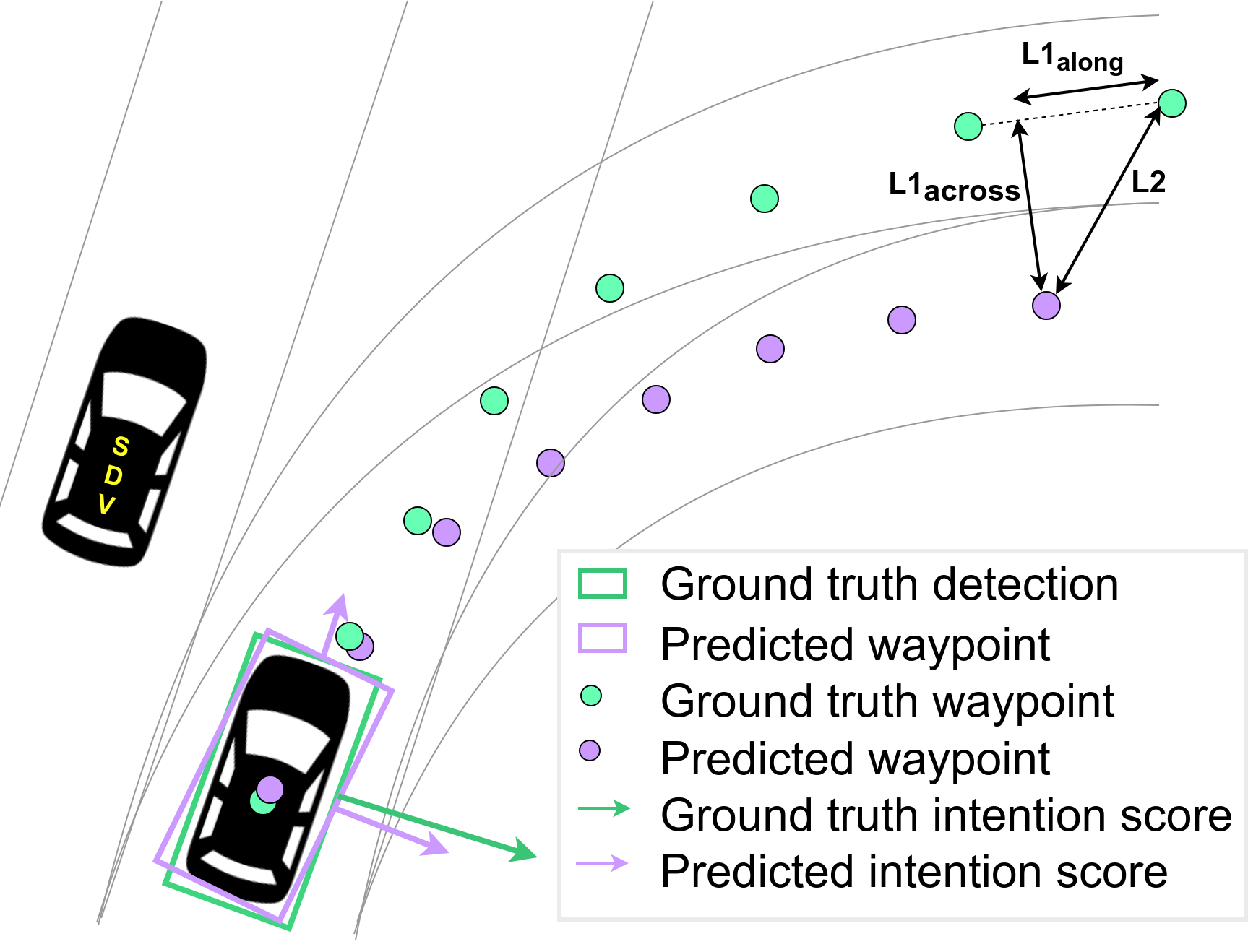}
    \newsubcap{Output parametrization}
    \label{fig:output_param}
  \end{subfigure}
  \begin{subfigure}[t]{.30\textwidth}
    \centering
    \includegraphics[width=\linewidth]{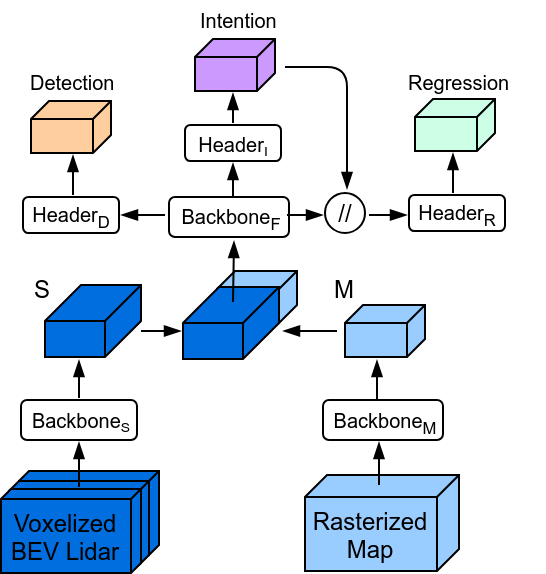}
    \newsubcap{Architecture overview}
    \label{fig:model_overview}
  \end{subfigure}
  \begin{subfigure}[t]{.30\textwidth}
    \centering
    \includegraphics[width=\linewidth]{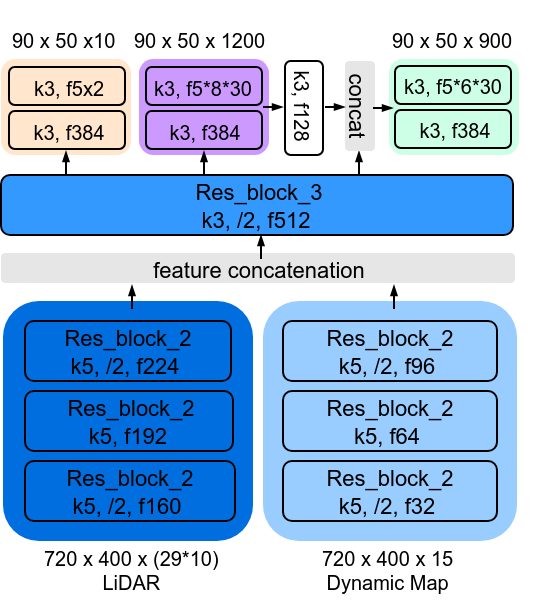}
    \newsubcap{Layer details}
    \label{fig:model_details}
  \end{subfigure}
\end{figure}

Work across different domains \cite{zhang2015sensor, snoek2005early, lewis2004sensor} has shown  that  late fusion  delivers stronger performance than early fusion. IntentNet exploits a late fusion of LiDAR and map information through an architecture consisting of a two-stream backbone network and three task-specific branches on top (see Figs.~\ref{fig:model_overview} and  \ref{fig:model_details}).

\paragraph{Backbone network:}
Our single-stage model takes two 3D tensors as input: the voxelized BEV LiDAR  and  the rasterization of our dynamic maps. 
We utilize a two-stream backbone network, where two different 2D CNNs process each data stream separately. %
The feature maps obtained from those subcomponents are then concatenated along the depth dimension and fed to the fusion subnetwork. %
We use a small downsampling coefficient in our network of 8x since each vehicle represents a small set of pixels in BEV, e.g., when using a resolution of 0.2 m/pixel, a car on average occupies  18 x 8 pixels.
To provide accurate long term intention prediction and motion forecasting, the network  needs to extract  rich motion information from the past and geometric details of the scene together with traffic  rule information. %
Note that vehicles typically drive at 50 km/h in urban scenes, traversing 42 meters in only 3 seconds. Thus we need our network to have  a sufficiently large effective receptive field \cite{luo2016understanding} to extract the desired information.   To keep both coarse and fine grained features, we exploit  residual connections \cite{he2016deep}.
We refer the reader to Fig.~\ref{fig:model_overview} for more details of our network architecture.

\paragraph{Header network:}
The header network is composed of three task specific branches that take as input the shared features from the backbone network. The detection branch outputs two scores for each anchor box at each feature map location, one for vehicle and one for background. An anchor is a predefined bounding box with orientation that serves as a prior for detection. Similar to \cite{faf, ren2015faster, liu2016ssd}, we use multiple anchors for each feature map location.
The intention network performs a multi-class classification over the set of high level actions, assigning a calibrated probability to the 8 possible behaviors at each feature map location.
The discrete intention scores are in turn fed into an embedding convolutional layer to provide extra features to condition the motion estimation. The motion estimation branch receives the concatenation of the shared features and the embedding from the high level action scores, and outputs the predicted trajectories for each anchor box at each feature map location.

\subsection{Learning} \label{learning}

Our model is fully differentiable and  thus can be trained end-to-end through back-propagation. In particular, we minimize a multi-task loss containing a regression term for the trajectory prediction over $T$ time steps, a binary classification term for the detection (background vs vehicle) and a multi-class classification for discrete intention. Thus
\[\mathcal{L}(\theta) = \mathcal{L}_{cla}(\theta) + \alpha \cdot \sum_{t=0}^{T} \lambda ^ t \cdot \mathcal{L}_{int}^t(\theta) + \beta \cdot \sum_{t=0}^{T} \lambda ^ t \cdot \mathcal{L}_{reg}^t(\theta)\]
where $t=0$ is the current frame and $t>0$ the future, $\theta$ the model parameters and $\lambda$ a temporal discount factor to ensure distance times into the future   do not dominate the loss as they are more difficult to predict. 
We mow define the loss functions we employ in more details. 

\paragraph{Detection:} We define a binary focal loss \cite{lin2017focal}, computed over all feature map locations and predefined anchor boxes, assigned using the matching strategy proposed in \cite{liu2016ssd}:
\[\begin{array}{ll}
\mathcal{L}_{cla}(\theta) = \sum_{i,j,k} -(1-\overline{p}_{i,j,k;\theta}) \cdot \log\overline{p}_{i,j,k;\theta}, \quad&
\overline{p}_{i,j,k;\theta} =
     \begin{cases}
       p_{i,j,k;\theta} & \quad \text{if} \quad q_{i,j,k} = 1, \\
       1 - p_{i,j,k;\theta} & \quad \text{otherwise}
     \end{cases}
\end{array}\]
where $i, j$ are the location indices on the feature map and $k$ is the index over the predefined set of anchor boxes; $q_{i,j,k}$ is the class true label and $p_{i,j,k;\theta}$  the predicted probability.
We define as positive samples, i.e. $q_{i,j,k} = 1$, those predefined anchor boxes having an associated ground truth box. In particular, for each anchor box, we find the ground truth box with the biggest intersection over union (IoU). If the IoU is bigger than a threshold of 0.5, we assign 1 to its corresponding label $q_{i,j,k}$. In case there is a ground truth box that has not been assigned to any anchor box, we assign it to the highest overlapping anchor box ignoring the threshold. Due to the imbalance of positive and negative samples we have found it helpful to not only use focal loss but also to apply hard negative mining during training. Thus, we rank all negative samples by their predicted score $p_{i,j,k}$ and take the top negative samples with a ratio of 3:1 with respect to the number of positive samples.

\paragraph{Trajectory regression:} We frame the detector regression targets as a function of the dimensions of their associated anchor boxes. As we reason in BEV, all objects have similar size as no perspective effect is involved. Thus, we can exploit object shape priors and make anchor boxes similar to real object sizes. This helps reduce the variance of the regression targets, leading to better training. In particular, we define
\[\begin{array}{lll}
\overline{c_x^t} = \frac{c_x^t-c_x^{anchor}}{w^{anchor}} \quad&
\overline{\phi_{sin}^t} = \sin\phi^t \quad&
\overline{w} = \log\frac{w}{w^{anchor}} \\\\
\overline{c_y^t} = \frac{c_y^t-c_y^{anchor}}{h^{anchor}} \quad&
\overline{\phi_{cos}^t} = \cos\phi^t \quad&
\overline{h} = \log\frac{h}{h^{anchor}}
\end{array}\]

We apply a weighted smooth L1 loss to the regression targets associated to the positive samples only. Note that for the future time steps ($t \, \epsilon \,[1, T - 1]$), the target set $\mathcal{R}_t$ does not include the bounding box size ($\overline{w}$, $\overline{h}$), which are only predicted at the current time step ($t=0$).
\[\begin{array}{ll}
\mathcal{L}_{reg}^t(\theta) = \sum\limits_{r \epsilon \mathcal{R}_t} \chi_r \cdot l_{r;\theta}^t, \quad&
l_{r;\theta}^t =
\begin{cases}
0.5 \cdot (x_{r;\theta}^t - y_r^t)^2, & \text{if } |x_{r;\theta}^t - y_r^t| < 1 \\
|x_{r;\theta}^t - y_r^t| - 0.5, & \text{otherwise }
\end{cases}
\end{array}\]
where $t$ is the prediction time step, $x_{r;\theta}^t$ refers to the predicted value of the $r$-th regression target, $y_r^t$ is the ground truth value of such regression target, and $\chi_r$ is the weight assigned to the r-th regression target.

\paragraph{Intention prediction:} We employ a cross entropy loss  over the set of high level actions. To address the high imbalance in the intention distribution, we downsample the dominant classes \textit{\big\{keep lane, stopping/stopped and parked\big\}} by 95\%. We found this strategy to work better than re-weighting the loss by the inverse frequency  in the training set. Note that we do not discard those examples for detection and trajectory regression.

\subsection{Inference} 

During inference, IntentNet  produces 3D detections with their respective continuous and discrete intentions for all vehicles in the scene in a single forward pass. Preliminary detections are extracted by applying a threshold of 0.1 to the classification probabilities, with the intention of achieving high recall. From these feature map locations, we examine the regression targets and anchor boxes, and use NMS to de-duplicate detections. Since our model can predict future waypoints for each vehicle, it provides a strong prior for associating detections among different time steps.
At any time step, we have a detection from the current forward pass and predictions produced at previous time steps. Therefore, by comparing the current location against past predictions of the future, we decode tracklets for each vehicle in the scene. This simple tracking system proposed in FaF \cite{faf} also allows us to recover missing false negatives and discard false positives by updating the classification scores based on previous predictions.

\begin{figure}
\begin{floatrow}
\capbtabbox{%
  \small
  \begin{tabular}{c ccccc}
\toprule
Model  				&  \multicolumn{5}{c}{Detection mAP @ IOU}  \\
{}  				&   0.5 	&  0.6 		&  0.7 		&  0.8 		&  0.9	\\
\midrule
SqueezeNet  	&   74.0 	&  62.3 	&  41.9 	&  13.8 	&  0.2	\\
SSD   				&   84.0	&  75.1 	&  58.2		&  26.0 	&  1.0	\\
MobileNet   		&   86.1	&  78.3		&  60.4 	&  27.5		&  1.1	\\
FaF   				&   89.8 	&  82.5		&  68.1		&  35.8 	&  2.5	\\
FaF'   				&   88.4 	&  80.1		&  64.1 	&  30.9		&  1.6	\\ 
\midrule[0.4pt]
IntentNet			&   \textbf{94.4} 	&  \textbf{89.4}		&  \textbf{75.4}		&  \textbf{43.5}	&  \textbf{3.9}	\\
\bottomrule
\end{tabular}

}{%
  \caption{Detection performance with objects containing \textit{p} $\geq$ 1 LiDAR points}%
  \label{table:det}
}
\ffigbox{%
  \includegraphics[width=\linewidth]{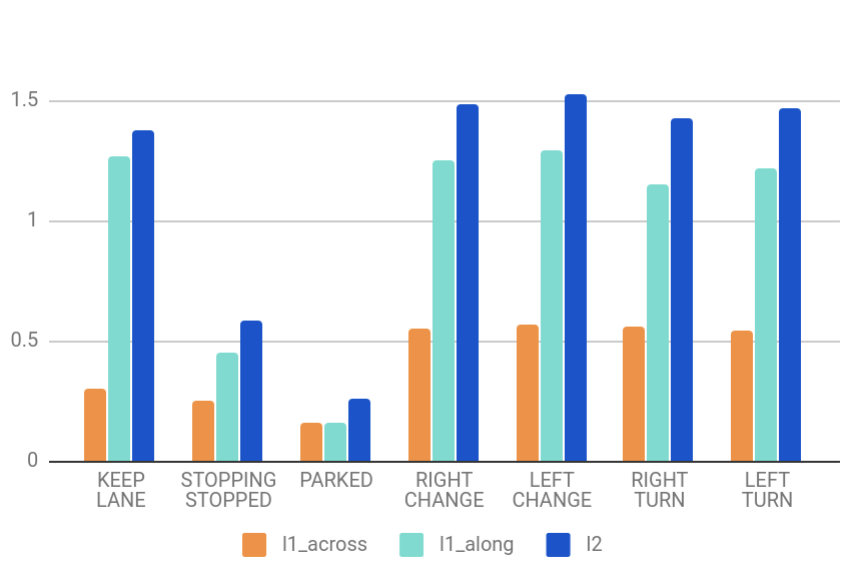}
}{%
  \vspace{-0.7cm}
  \caption{L1 error along and across track and L2 error split by ground truth discrete intention}%
  \label{fig:reg_chart}
}
\end{floatrow}
\end{figure}
\section{Experimental Evaluation}
We evaluate our approach in the tasks of detection, intention prediction and motion forecasting. 
To enable this, we collected a large scale dataset from a roof-mounted LiDAR on top of a self-driving vehicle driving around several cities in North America. It contains over 1 million frames collected from over 5,000 different scenarios, which are sequences of 250 frames captured sequentially at 10 Hz. 
Our labels are tracklets of 3D non-axis align bounding boxes. 
The dataset is highly imbalanced and thus challenging, containing the following number of examples of each behavior; \textit{keep lane}: 10805k, \textit{turn left}: 546k, \textit{turn right}: 483k, \textit{left change lane}: 290k, \textit{right change lane}: 224k, \textit{stopping/stopped}: 12766k, \textit{parked}: 29110k and \textit{others}: 100k.

\paragraph{Implementation details:}

We use a birds eye view (BEV) region with $L = 144$, $W = 80$ meters from the center of the autonomous vehicle and $H = 5.8$ meters from the ground, for both training and evaluation. We set the  resolution to be $\Delta L = \Delta W = \Delta H = 0.2$ meters. We use $T = 10$ past LiDAR sweeps to provide enough context in order to perform accurate long term prediction of 3 seconds. 
Thus, our input is a 3D tensor of shape $(29 \cdot 10,720, 400)$. We use 5 predefined anchor boxes of size 3.2 meters and $1:1, 1:2, 2:1, 1:6, 6:1$, and a threshold of 0.1 for the classification score to consider a detection positive during inference. We train our model from scratch using Adam optimizer \cite{kingma2014adam}, a learning rate of 1e-4 and a weight decay of 1e-4. We employ a temporal discount factor $\lambda = 0.97$   for our future predictions.  We used a batch of size 6 for each GPU and perform distributed training on 8 Nvidia 1080 GPUs for around 24h. %

\paragraph{Detection results:} We compare mean average precision (mAP) at different IoU levels with other object detectors that can also run inference in real-time (i.e., less  than 100ms) including SqueezeNet \cite{iandola2016squeezenet}, SSD \cite{liu2016ssd}, MobileNet \cite{howard2017mobilenets}, FaF \cite{faf} and FaF' (adaptation of FaF where the motion forecasting header is trivially extended to predict 3 seconds). As shown in Table \ref{table:det},  our model is clearly superior  across all IoU levels. Note that all models use the same anchor boxes and downsampling factor for the comparison to be consistent.

\begin{table}[t]
\centering
\small
\begin{tabular}{c cccc cccc cccc}
\toprule
Model  	&  \multicolumn{4}{c}{L1 error along track (m)}  &  \multicolumn{4}{c}{L1 error across track (m)} & \multicolumn{4}{c}{L1 error heading (deg)} 	\\
{}  				&   0s 	&  1s 	&  2s 	&  3s 			&   0s 	&  1s 	&  2s 	&  3s							&   0s 	&  1s 	&  2s 	&  3s			\\
\midrule
FaF   				&  0.28 &  0.53 &  - 	&  - 			&  0.17	& 0.26	& -		& -								&  6.06	& 6.47	& -		& -			\\
FaF'   				&  0.29 &  0.49 &  0.95 &  1.67 		&  0.21	& 0.29	& 0.45	& 0.69 							&  6.35	& 6.57	& 7.16	& 8.02			\\ 
\midrule[0.4pt]
IntentNet*  	&   0.27 & 0.47 &  0.92 &  1.64 		&  0.18	& 0.27	& 0.43	& 0.65 							&  5.81	& 6.09	& 6.69	& 7.62			\\ 
IntentNet   		&   \textbf{0.26} & \textbf{0.46}	& \textbf{ 0.91} &  \textbf{1.61 }	&  \textbf{0.15	}& \textbf{0.21}	& \textbf{0.34}	& \textbf{0.53} &   \textbf{5.14}	& \textbf{5.35}	& \textbf{5.83}	& \textbf{6.54}			\\ 
\bottomrule
\end{tabular}
\caption{Regression error computed over the intersection of true positive detections from the four models. IntentNet* is our model without map and without high level actions}
\label{table:reg}
\end{table}

\paragraph{Trajectory regression results:} To the best of our knowledge, FaF \cite{faf} is the only previous work that performs joint detection and motion forecasting from LiDAR. 
As shown in Table \ref{table:reg} IntentNet outperforms both FaF and FaF' in both along and across track errors as well as heading. The performance leap is a combination of network architecture improvements and the usage of prior knowledge in the form of maps. To be fair, the metrics are reported over the intersection of true positives of all the  four models, which represents over 90 \% of the validation set. 
In addition,  Fig.  \ref{fig:reg_chart} depicts   regression metrics split by high level action. We highlight that IntentNet is able to learn complex velocity profiles such as the ones in turns and lane changes, keeping the along track error almost as low as in lane keeping.

\paragraph{Intention prediction:}
We compare our approach with the models proposed in \cite{phillips2017generalizable}, adapting their classifiers to perform our task, which is a generalization of theirs. Phillips \textit{et al.} extract several handcrafted features including \textit{base features} such as velocity, acceleration, number of lanes to the curb and the median and headway distance to preceding vehicle, at the current time step; \textit{history features} containing such information for previous time steps; \textit{traffic features} containing up to six neighbours base features and  \textit{rule features} such as whether the car can turn left/right from its lane. We then train their model in our dataset. 
Table \ref{table:highlvl} shows the results of their two best performers models: an MLP and an LSTM. Our model clearly outperforms the baselines, especially on the less represented high level actions where the baselines are  unable to recognize turns or lane changes. Note that we also tried applying the same downsampling method we use to the baselines but it resulted in worse overall performance and therefore was omitted here. As shown, IntentNet is able to generalize despite the extreme imbalance in the behavior distribution. %

\begin{table}[t]
\centering
\small
\begin{tabular}{ccccccccccc}
\toprule
Model  		& Metric 	&  P		& 	S		& 	LK		&	TR 		& TL	& LCR	& LCL	& Others 	& Mean \\
\midrule
MLP			&   Acc		&   63.4 	&  68.5 	&  79.4 	&	98.8 	& 98.6	& 100	& 100	& 100 		& 88.6 \\
{}			&   F1		&   67.2 	&  39.0 	&  32.5 	&	00.0 	& 00.0	& 00.0	& 00.0	& 00.0 		& 17.3 \\
\midrule[0.2pt]
LSTM		&   Acc		&   66.9 	&  69.2 	&  79.9 	&	99.2	& 98.7	& 100	& 100	& 100 		& 89.3 \\
{}			&   F1		&   69.8 	&  48.1 	&  37.8 	&	00.0 	& 00.0	& 00.0	& 00.0	& 00.0 		& 19.5 \\
\midrule[0.4pt]
IntentNet 	&   Acc		&   93.9 	&  89.5 	&  89.5 	&	97.5 	& 96.6	& 98.3	& 97.9	& 97.5 		& 95.1 \\
w/o map		&   F1		&   98.7	&  80.4		&  80.4		&	51.9	& 52.1	& 18.2	& 24.1	& 21.6 		& 53.4 \\
\midrule[0.4pt]
IntentNet	&   Acc		&  \textbf{ 99.2}	&  \textbf{94.9}		&  \textbf{93.5}		&	\textbf{98.6}	& \textbf{98.6}	& \textbf{99.2}	& \textbf{99.1}	& \textbf{98.5} 		& \textbf{97.7} \\
{}			&   F1		&   \textbf{98.8}	&  \textbf{91.0}		&  \textbf{88.7}		&	\textbf{66.4}	& \textbf{73.1}	& \textbf{50.3}	& \textbf{55.5}	& \textbf{45.2 }		& \textbf{71.1} \\
\bottomrule
\end{tabular}
\caption{Average intention prediction performance at the current timestep}
\label{table:highlvl}
\end{table}

\paragraph{Ablation study:} 
We conducted an ablation study in order to evaluate how much each of the contributions proposed in this paper helped towards achieving the final results, which is shown in Table \ref{table:ablation}. Using a 2D CNN with early fusion of the different LiDAR sweeps delivers a much robust detector, being able to understand the time dimension better than a 3D CNN as proposed in \cite{faf}. Increasing the context from 0.5 seconds to 1 second (5 to 10 input LiDAR sweeps) gives us a small gain, decreasing the long term L2 error. Notice that even though the detector is able to have higher recall while using the same confidence threshold (0.1), the regressions become better in average (even when taking into account those harder examples). Adding the loss for the discrete intention estimation degrades the detector/regressor performance due to the fact that it is very hard to predict behavior purely based on motion. However,  after adding the map,  the system is able to predict the high level behavior of vehicles and thus adding the loss improves general performance. For a direct comparison to FaF \cite{faf}, its detection results (mAP) are 89.8 (IOU=0.5,p$\geq$1), 93.3 (IOU=0.5,p$\geq$5), 68.1 (IOU=0.7,p$\geq$1) and 73.4 (IOU=0.7,p$\geq$5); its recall is 92.9\% and its L2@0s is 0.36 meters.

\begin{table}[t]
\centering
\small
\begin{tabular}{ccccccccccccc}
\toprule
\multicolumn{2}{c}{Conv type} 	& 	\multicolumn{2}{c}{Context} 	& 	 M		& 	I	& \multicolumn{2}{c}{mAP@0.5}  & \multicolumn{2}{c}{mAP@0.7} 	& Recall	& 	L2@0s   & L2@3s	\\
3D 	 		 & 		2D 			& 	0.5s  	& 1.0s				 	& 	 {}		& 	{}		& $p\geq1$ & $p\geq5$  		& $p\geq1$  & $p\geq5$ 			& {}  		& 	(m)    	& (m)	\\
\midrule
\cmark 	 	 &   	{}			&  \cmark 	&  {} 					&  	{} 		&  	{}		&  88.2	 	&  	92.0		&  	62.9	  & 68.3 		& 91.3			&   0.37 	& 2.16	\\
{}			 &   	\cmark 		&  \cmark 	&  {} 					&  	{} 		&  	{}		&  91.9 	&  	94.8 		&  	71.7	  & 76.8		& 93.7			&  	0.33 	& 1.90	\\
{}			 &   	\cmark 		&  {} 		&  \cmark 				&  	{} 		&  	{}		&  92.4 	&  	95.0 		&  	72.8	  & 77.5 		& 94.1			&  	0.33  	& 1.85	\\
{}		 	 &  	\cmark 		&  {} 		&  \cmark  				&  	{} 		&   \cmark	&  90.1		&  	92.5 		&  	69.8	  & 74.4 		& 93.8			&  	0.34	& 1.84	\\
{}		 	 &  	\cmark 		&  {} 		&  \cmark 				&  	\cmark 	&  	{}		&  94.3 	&  	\textbf{96.3} 		&  	\textbf{75.5}	  & \textbf{79.6}		& 95.8			&  	\textbf{0.33}  	& 1.80	\\
{}		 	 &  	\cmark 		&  {} 		&  \cmark  				&  	\cmark 	&   \cmark	&  \textbf{94.4} 	&  	96.2		&  	75.4	  & \textbf{79.6}		& \textbf{95.9}			&  	\textbf{0.33} 	& \textbf{1.79}	\\
\bottomrule
\end{tabular}
\caption{Ablation study of IntentNet different contributions. \textit{M} column states whether the model uses the map or not. \textit{I} column states if discrete intention is being predicted.}
\label{table:ablation}
\end{table}

\begin{figure}[t]
  \begin{subfigure}[t]{.325\textwidth}
    \centering
    \includegraphics[width=\linewidth]{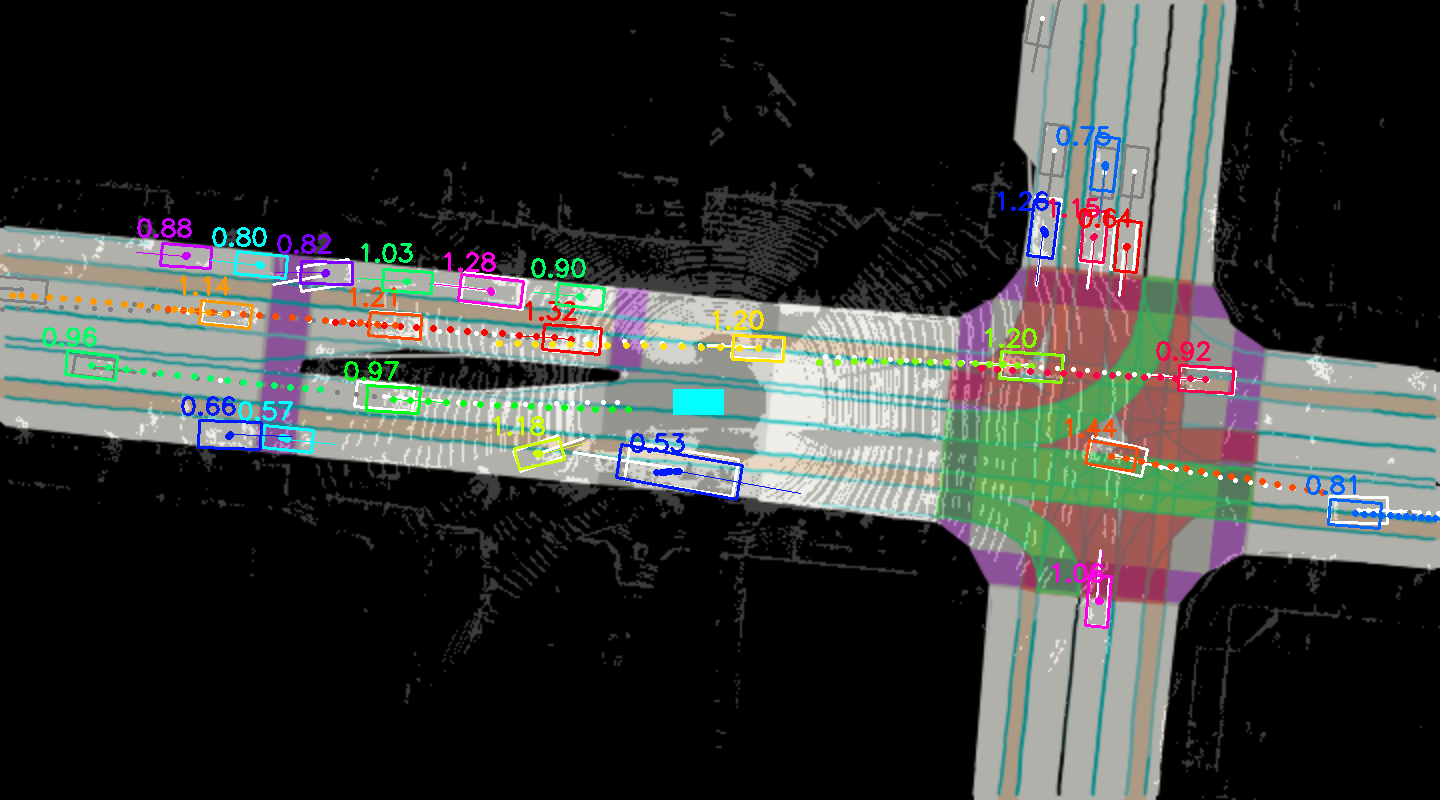}
  \end{subfigure}
  \begin{subfigure}[t]{.325\textwidth}
    \centering
    \includegraphics[width=\linewidth]{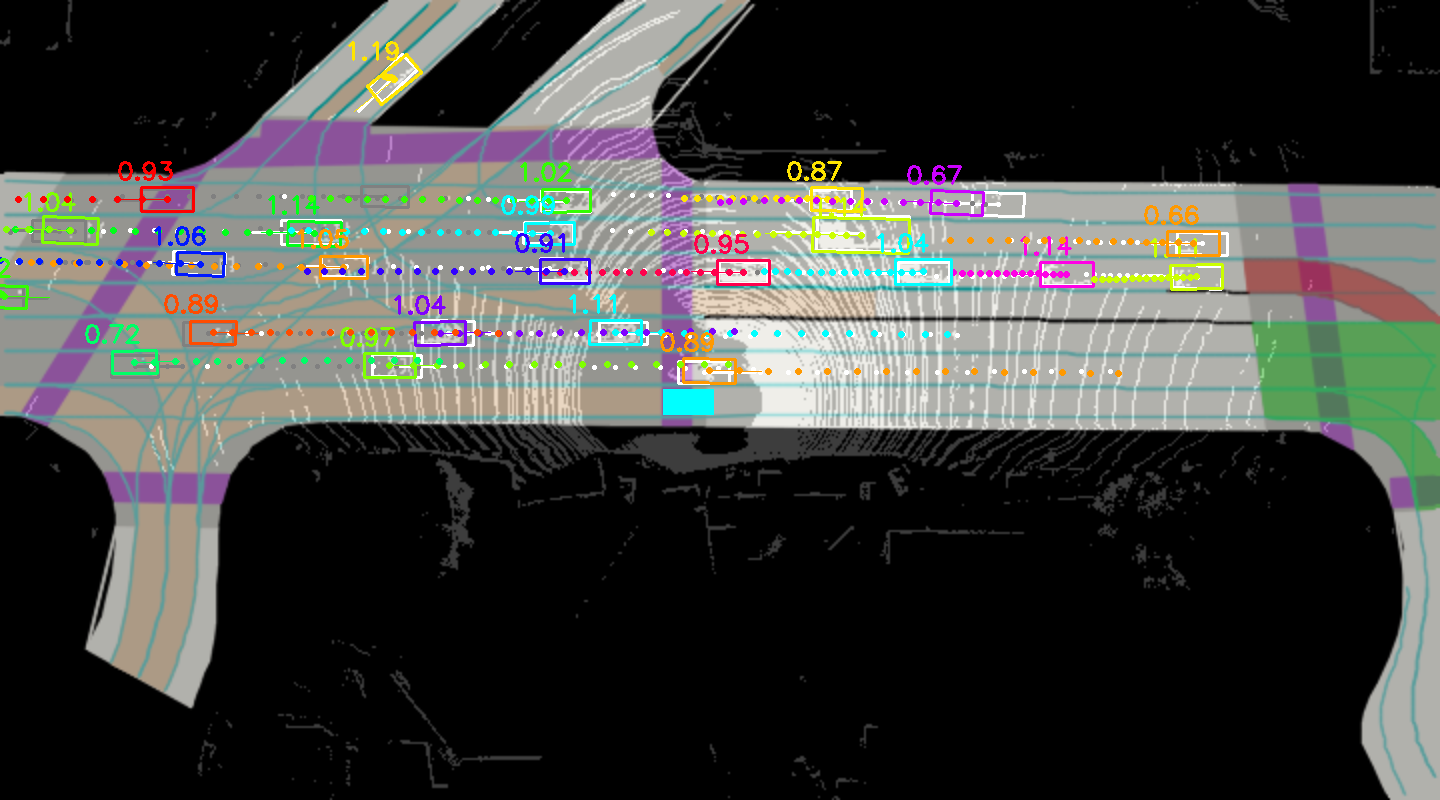}
  \end{subfigure}
  \begin{subfigure}[t]{.325\textwidth}
    \centering
    \includegraphics[width=\linewidth]{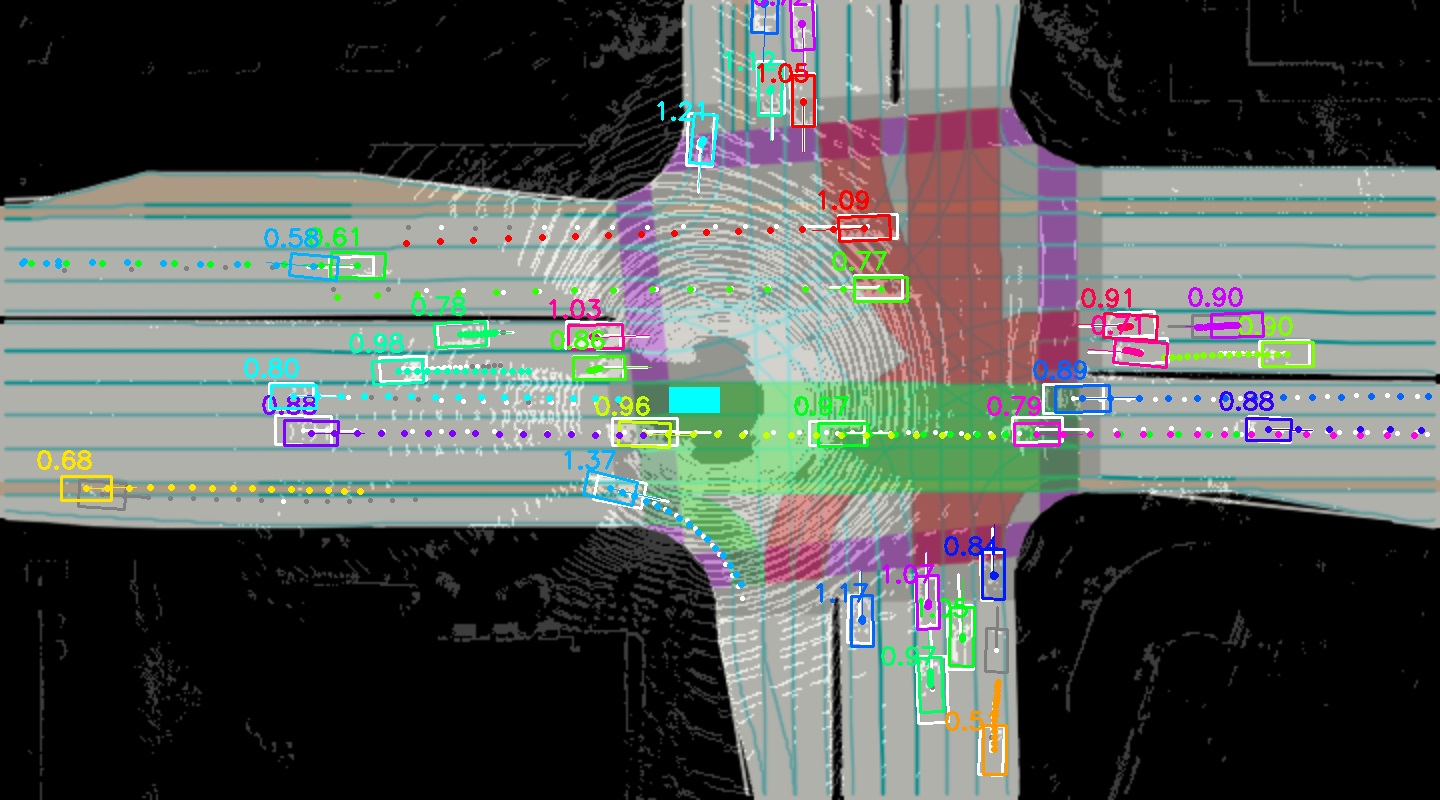}
  \end{subfigure}

  \begin{subfigure}[t]{.325\textwidth}
    \centering
    \includegraphics[width=\linewidth]{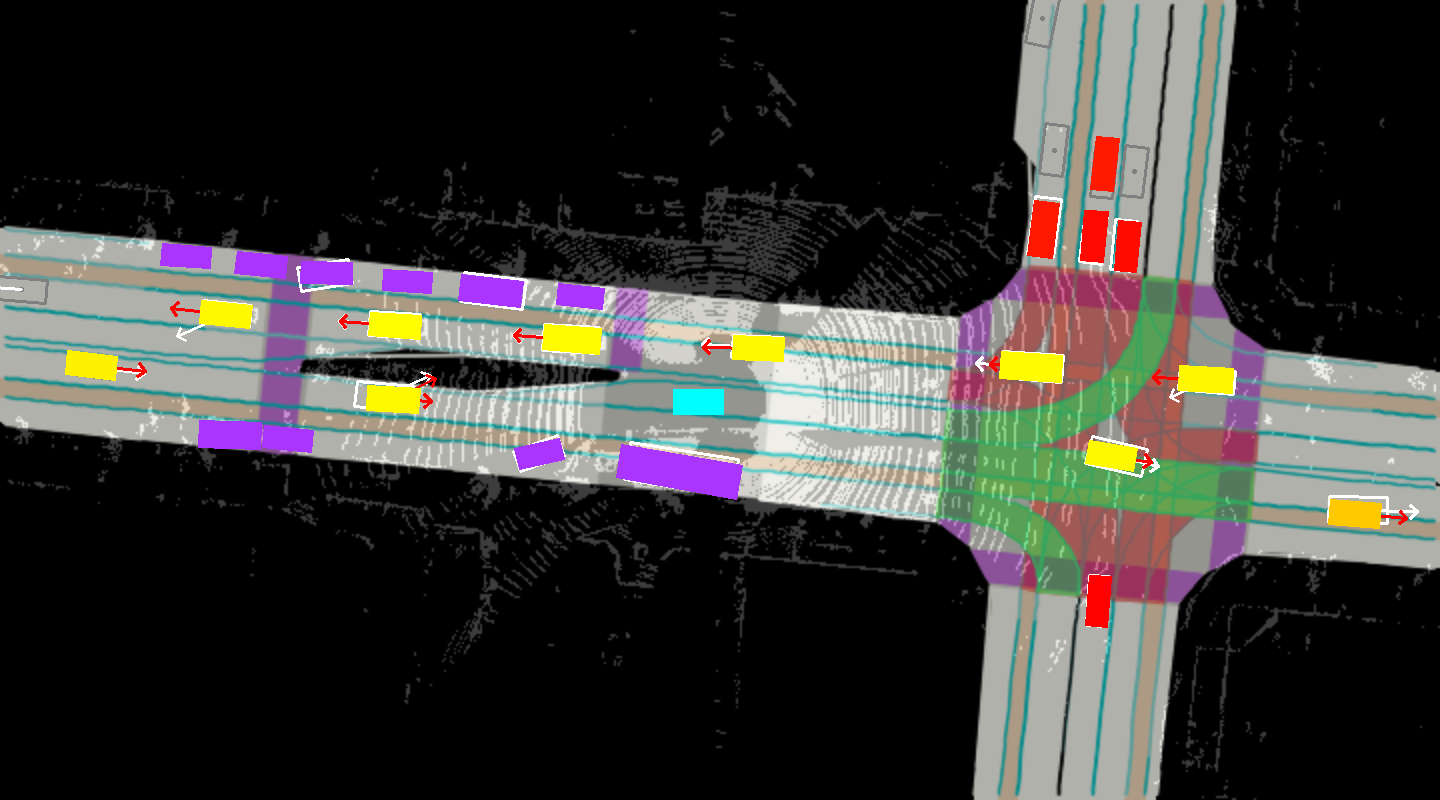}
  \end{subfigure}
  \begin{subfigure}[t]{.325\textwidth}
    \centering
    \includegraphics[width=\linewidth]{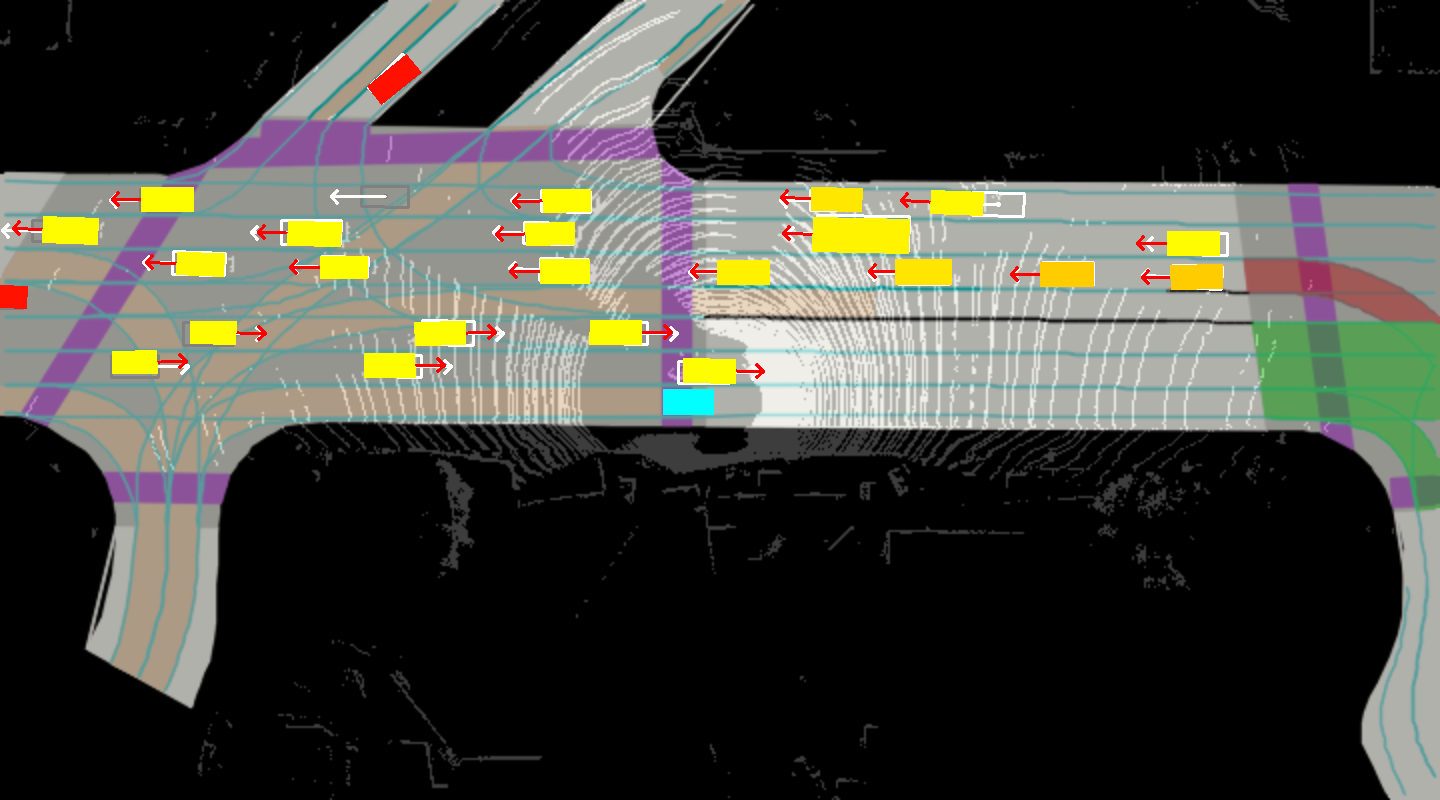}
  \end{subfigure}
  \begin{subfigure}[t]{.325\textwidth}
    \centering
    \includegraphics[width=\linewidth]{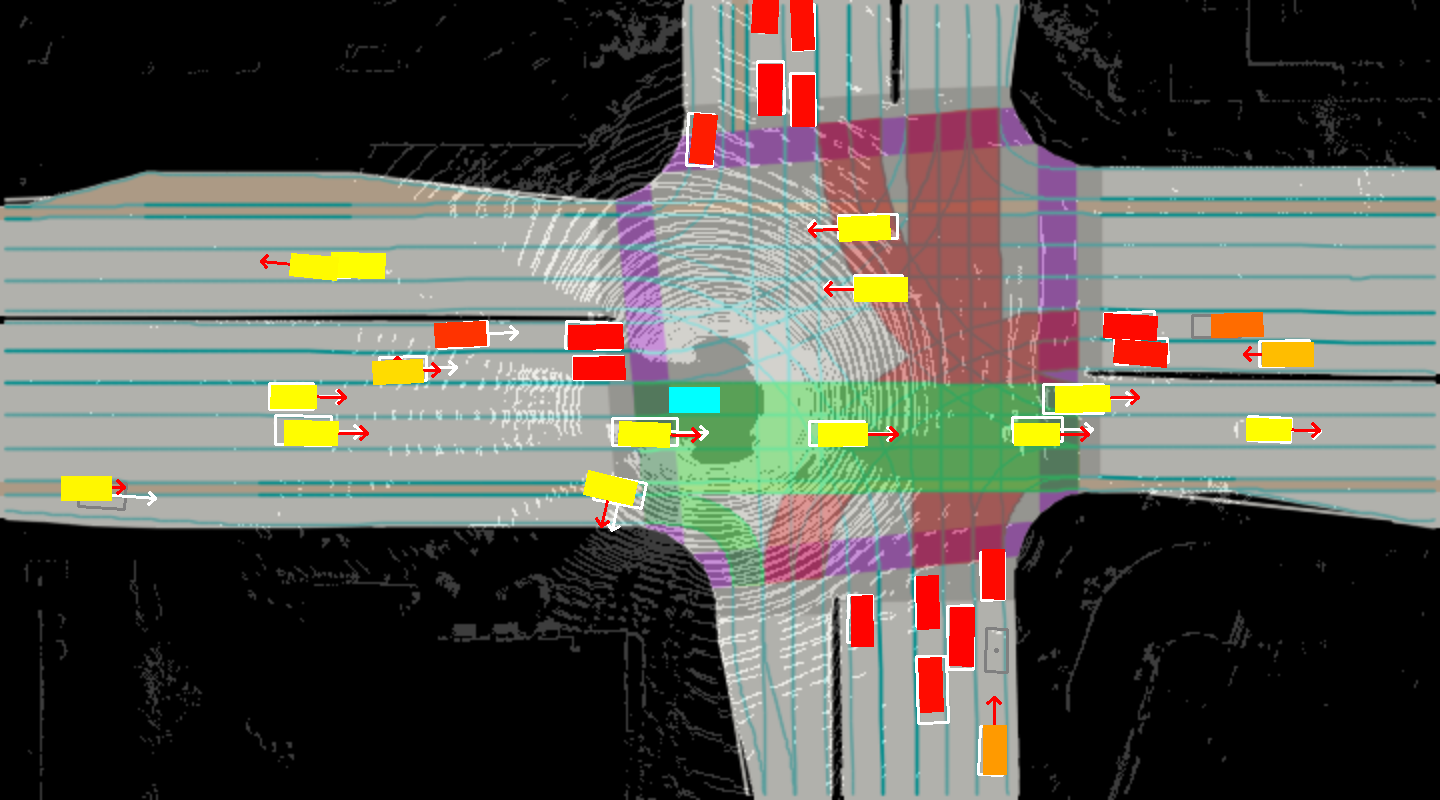}
  \end{subfigure}
  
\caption{Qualitative results. Ground truth is displayed in white (grey if the ground truth did not have LiDAR points) and predictions in color. Legend for high level actions - \textit{parked}: purple box, \textit{stopping/stopped}: red box, \textit{keep lane}: straight arrow, \textit{turns}: 90 $^{\circ}$ arrow, \textit{lane changes}: 30 $^{\circ}$ arrow. For the arrows, the longer the higher the probability. \textit{Others} class is not shown for simplicity.}
\label{fig:qualitative}
\end{figure}

\paragraph{Qualitative results:}
Fig.  \ref{fig:qualitative} shows our results in the 144 x 80 meters region. We can see 3 pairs of frames belonging to different scenarios. We display the detections and continuous intent prediction in the top row and discrete intent prediction in the bottom row. Our model is able to predict lane changes and detect big size vehicles (left), have a high precision and recall in cluttered scenes (center) and predict turns (right).

\section{Conclusion}
\label{sec:conclusion}

In this paper we introduce IntentNet, a learnable end-to-end model that is able to tackle the tasks of detection and intent prediction of vehicles in the context of  self-driving cars. By exploiting 3D point clouds produced by a LiDAR sensor and prior knowledge of the scene coming from an HD  map, we are able to achieve higher performance than previous work across all tasks, with a single neural network. 
In the future, we plan to investigate how more sophisticated algorithms can model the statistical dependencies between discrete and continuous intention. %
We also plan to extend our approach to handle pedestrians and bicyclists.

\clearpage

\bibliography{mybib}  %

\end{document}